\documentclass[runningheads]{llncs}

\usepackage{graphicx}
\usepackage[dvipsnames]{xcolor}
\usepackage{tikz}
\usepackage{pgfplots}

\usepackage{url}

\usepackage[textsize=footnotesize]{todonotes}
\usepackage{siunitx}

\title{
  Bold Hearts \\ Team Description for RoboCup 2019 \\ (Humanoid Kid Size League)
}
\titlerunning{Bold Hearts TDP 2019}

\author{Marcus M. Scheunemann \and
Sander G. van Dijk \and
Rebecca Miko \and
Daniel Barry \and
George M. Evans \and
Alessandra Rossi \and
Daniel Polani}
\authorrunning{MM. Scheunemann et al.}

\institute{School of Computer Science \\
University of Hertfordshire, Hatfield AL10 9AB, UK \\
\email{marcus@mms.ai}\\
\url{https://robocup.herts.ac.uk}}

\usepackage[
final 
]{hyperref}
\hypersetup{
 pdftitle={Bold Hearts Team Description for RoboCup 2019 (Humanoid Kid Size League},
 pdfsubject={An overview of recent developments for a successful participation in RoboCup 2019},
 pdfauthor={Marcus M. Scheunemann, Sander G. van Dijk, Rebecca Miko, Daniel Barry, George M. Evans, Daniel Polani},
 pdfkeywords={autonomous robots, robot hardware, RoboCup, simulator, ROS 2, deep learning, semantic segmentation}
}

\begin{document}

\maketitle

\begin{abstract}
We participated in the RoboCup 2018 competition in Montreal with our newly developed BoldBot based on the Darwin-OP and mostly self-printed custom parts.
This paper is about the lessons learnt from that competition and further developments for the RoboCup 2019 competition.
Firstly, we briefly introduce the team along with an overview of past achievements.
We then present a simple, standalone 2D~simulator we use for simplifying the entry for new members with making basic RoboCup concepts quickly accessible. 
We describe our approach for se\-man\-tic-seg\-men\-tation for our vision used in the 2018 competition, which replaced the lookup-table~(LUT) implementation we had before.
We also discuss the extra structural support we plan to add to the printed parts of the BoldBot and our transition to ROS 2 as our new middleware.
Lastly, we will present a collection of open-source contributions of our team. 
\end{abstract}

\section{Bold Hearts}
\label{sec:boldhearts}

The team Bold Hearts has been founded as part of the Adaptive Systems
Research Group at the University of Hertfordshire.
The team started participating in RoboCup in 2003 in the simulation
leagues and made a transition to the Humanoid KidSize League in 2013. We hope to participate in that league in 2019 for the seventh year in a row.

The following are the main achievements of team
Bold Hearts in the Humanoid League over the last few years.

\begin{itemize}
\item Quarter-finalist RoboCup World Championchip 2017 (1st in group) 
\item 2nd round RoboCup World Championship 2016 (1st in group)
\item 1st Iran Open 2016
\item 2nd round RoboCup World Championship 2015 (1st in group)
\item 3rd German Open 2015
\item 2nd RoboCup World Championship 2014
\end{itemize}

\section{Introducing New Members Gradually}
Recruiting new members is a crucial task, as with most RoboCup teams. 
It is important for the team's overall success to continuously recruit new members and transfer knowledge to the new generations. 

We have always kept a well maintained wiki for new members to read up on the given infrastructure. 
Additionally, we have made it easier to set up the code by using tools such as Ansible and Docker\footnote{\url{https://www.ansible.com/} and \url{https://www.docker.com/}}. 
However, there has been a lack of students with C++ knowledge. 
This is the language of choice for our custom framework, which is presented in previous team description papers~\cite{tdp-18}.
Additionally, robotics itself is very complex, therefore working with real robots without pre-knowledge is a challenge in itself. 

We decided to tackle this issue on different levels. 
Firstly, we plan on moving to a framework which allows modular development in Python and C++, see section~\ref{sec:middleware} for more details.

Students then need to understand the level of complexity of robotic tasks. 
Some known contributors to the RoboCup community approached this issue gradually themselves, by firstly participating in, e.g., simulation league and only later entering the hardware league.
We want to emulate this locally, by offering students a very simple and accessible idea of RoboCup with a simple, standalone 2D simulator called PythoCup. It is written in Python and was firstly developed for the Humanoid soccer school 2013 by our team. 
It has now been adapted for the use of PyGame and is published on GitLab\footnote{\url{https://gitlab.com/boldhearts/pythocup}}.

\begin{figure}[ht]
  \centering
  \includegraphics[width=.49\textwidth]{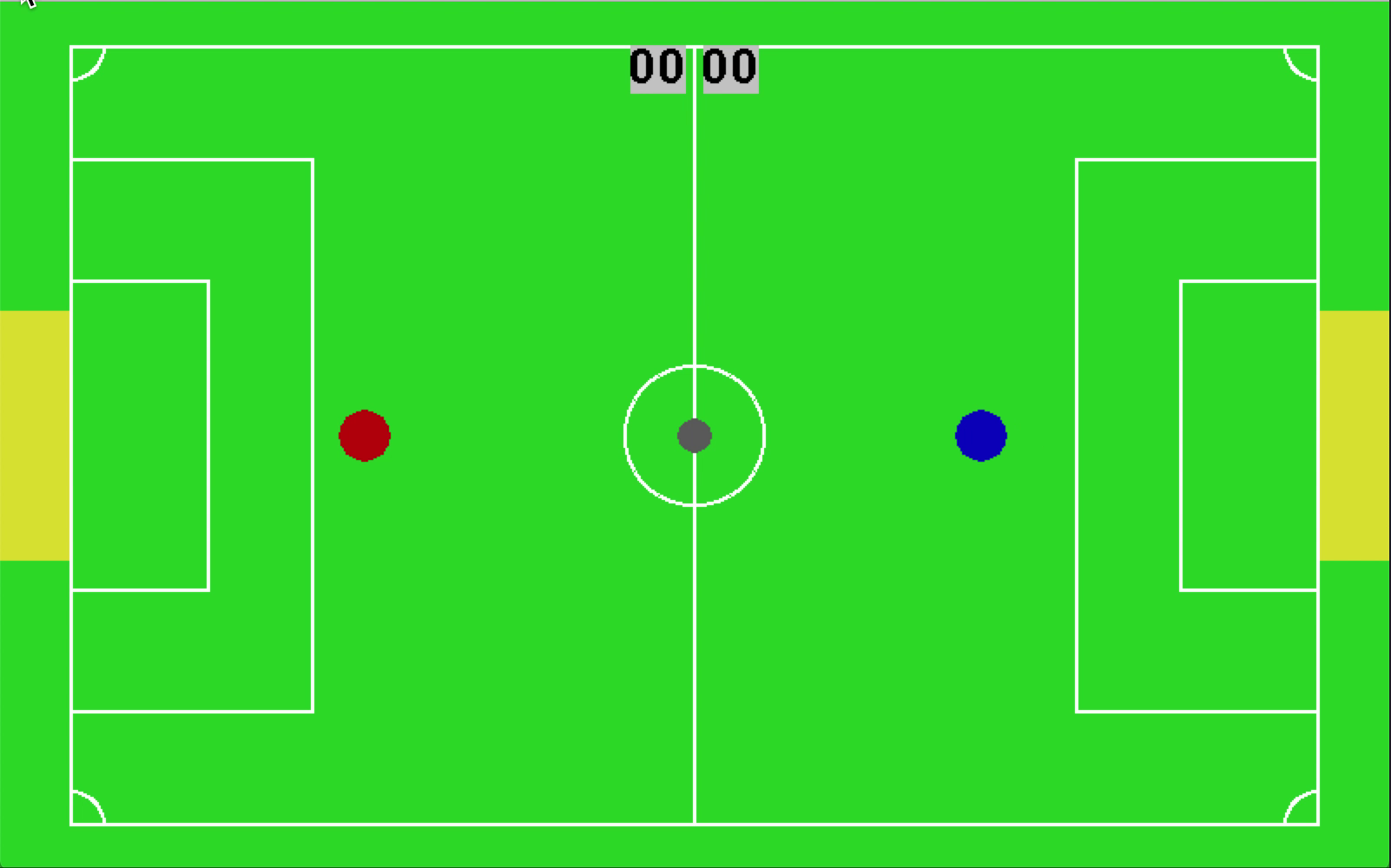}
  \includegraphics[width=.49\textwidth]{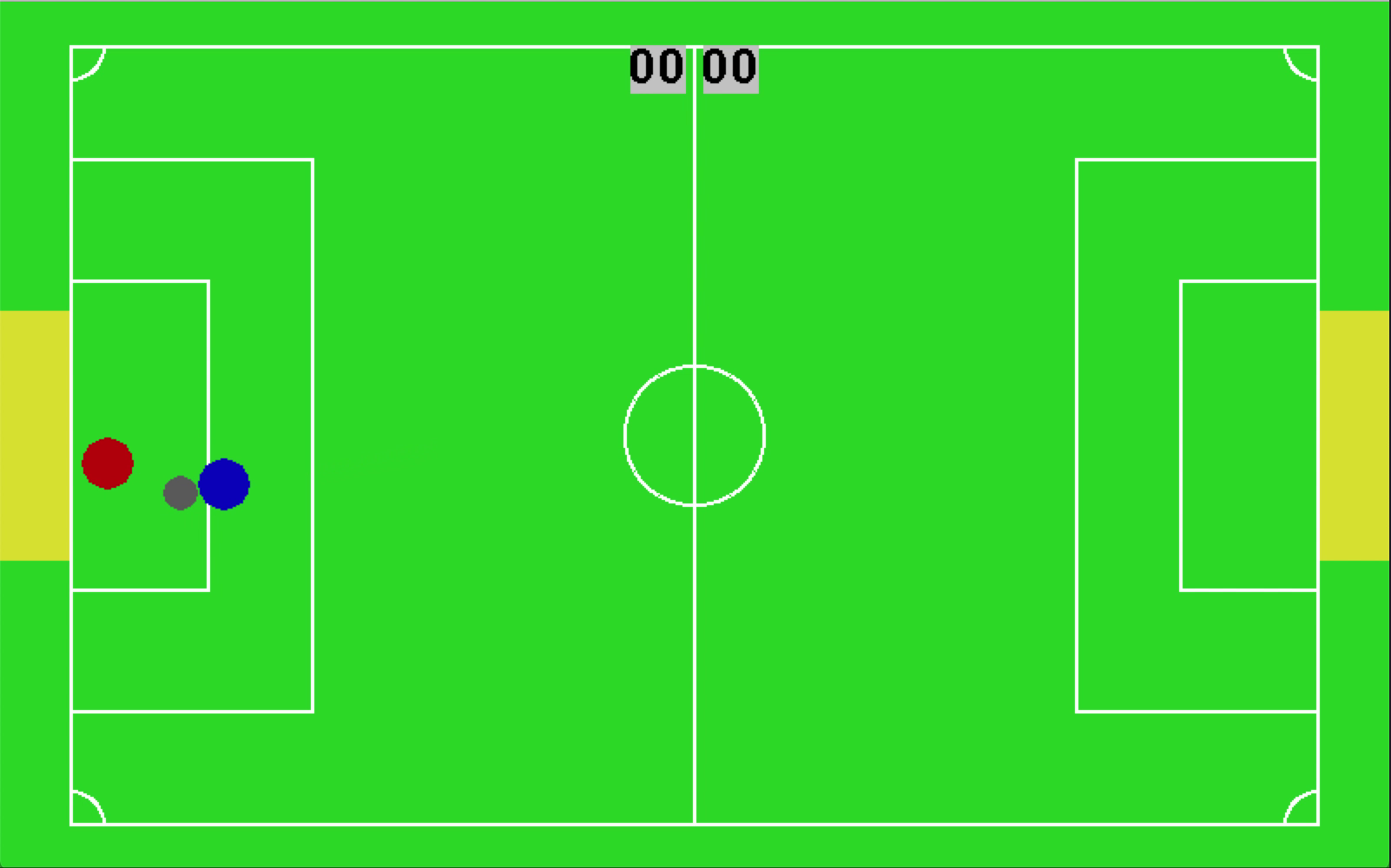}
  \caption{Two scenes of a PythoCup game. The left screenshot shows the moment before the game starts. The right screenshot shows the blue player attacking the goal of the red player.}
  \label{fig:pythocup_screenshot}
\end{figure}%

We expect several benefits for new students/participants and existing team members. Setting up PythoCup is simple and manipulating the behaviour of the robots can be achieved in a few steps, yet it offers the possibility to achieve already quite sophisticated agents. 
We hope that this will ease the process for new members joining the team. 
Additionally, we expect that some important robotic related problems will derive naturally and therefore we hope that new members get a glimpse of RoboCup related problems quickly.
Another benefit is that those who have learnt these skills can then help introduce new members, creating a pyramid of experience.

After mastering PythoCup, the next step will be to allow students to set up our code for the humanoid robots. 
They will be given an isolated modular problem to solve. 
Testing will be done in the simulator (e.g.~Gazebo) and already small changes will yield different, visible output.

\section{Robotic Hardware and Design}
\begin{figure}[htb]
  \centering
  \includegraphics[width=.93\textwidth]{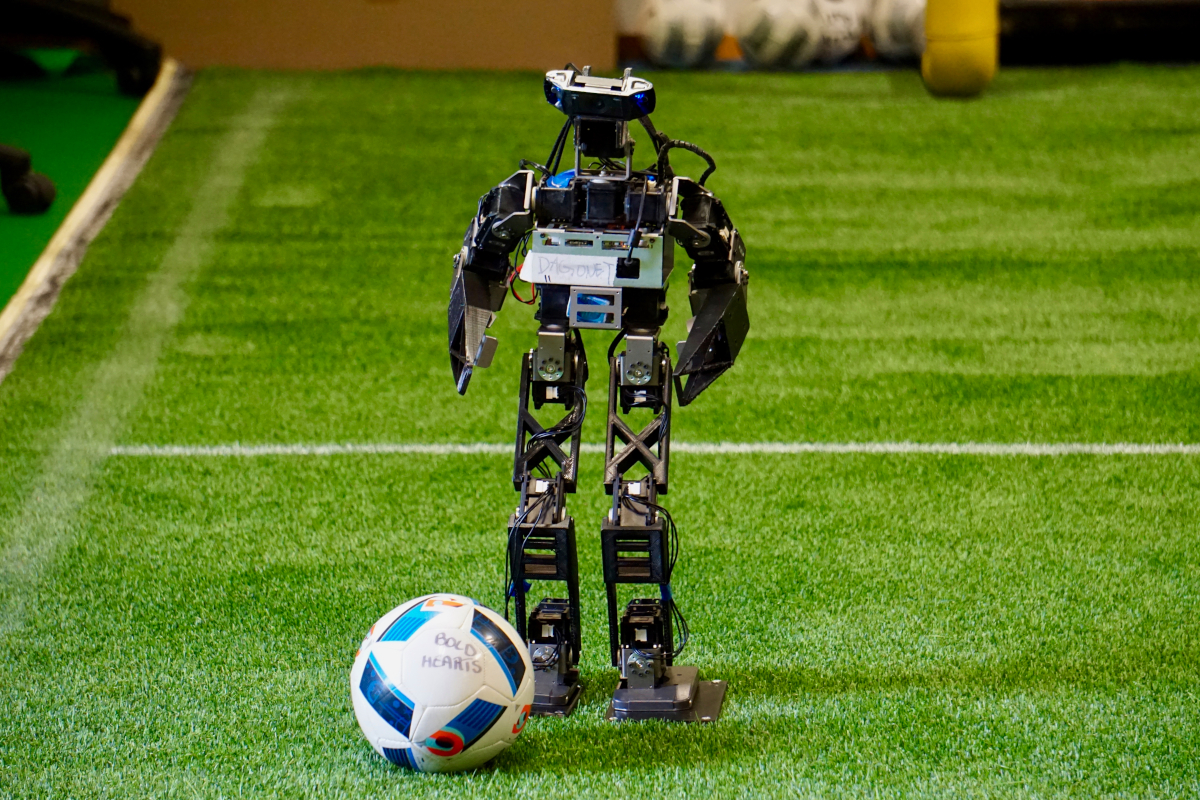}
  \caption{The BoldBot robot in its second version. It is incrementally developed with a Darwin-OP as its base. The torso has been scaled up to fit an Odroid-XU4, the new main processing unit. The shin, thigh, arm, head bracket and the foot plate have been redesigned, scaled up and 3D~printed.}
  \label{fig:boldbot}
\end{figure}
As described in our last years' team description paper, we started with incrementally developing a new robot platform based on the Darwin-OP~\cite{tdp-18}.

The main processing unit has been replaced with an Odroid-XU4. Shins, legs, foot plates, head bracket and arms have been redesigned and 3D printed. Figure~\ref{fig:boldbot} shows one of our robots with its newly designed parts. At the RoboCup 2018 competition, we participated with 4 robots of that configuration. The robots were equipped with four different webcam models: Logitech C910, C920, C920c and C930e.
For this year's configuration, we decided on using Logitech C920 Pro HD webcam for all robots.

For the self-printed limbs, we mainly used PLA and ABS and a range of different printers. PLA seems to be most sturdy and well suited for our needs. One of our biggest challenges during the RoboCup 2018 competition was mounting the printed parts to the servos.
The plastic parts had to resist a lot of stress on a small area of contact of the screws. When the plastic parts broke, they usually did so around the horn mounting area.

\begin{figure}[ht]
  \centering
  \includegraphics[width=0.3565\textwidth]{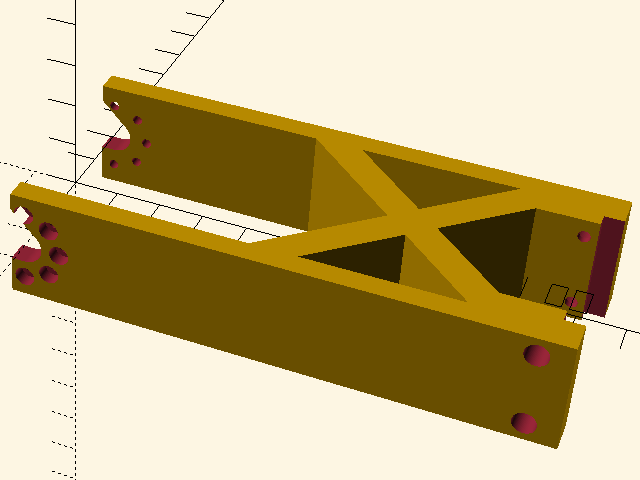}
  \includegraphics[width=0.3565\textwidth]{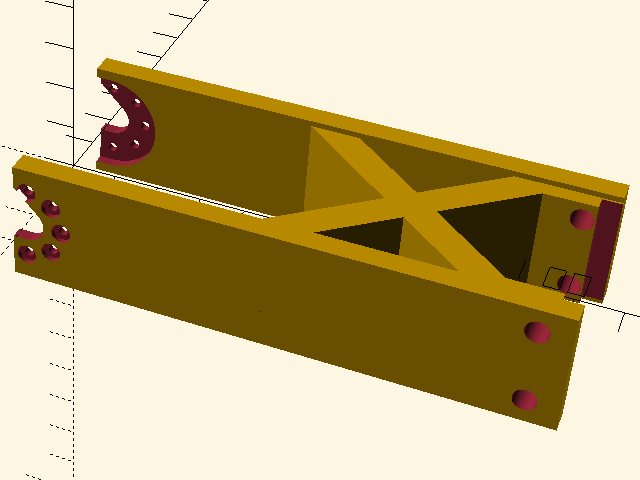}
  \includegraphics[width=0.271\textwidth]{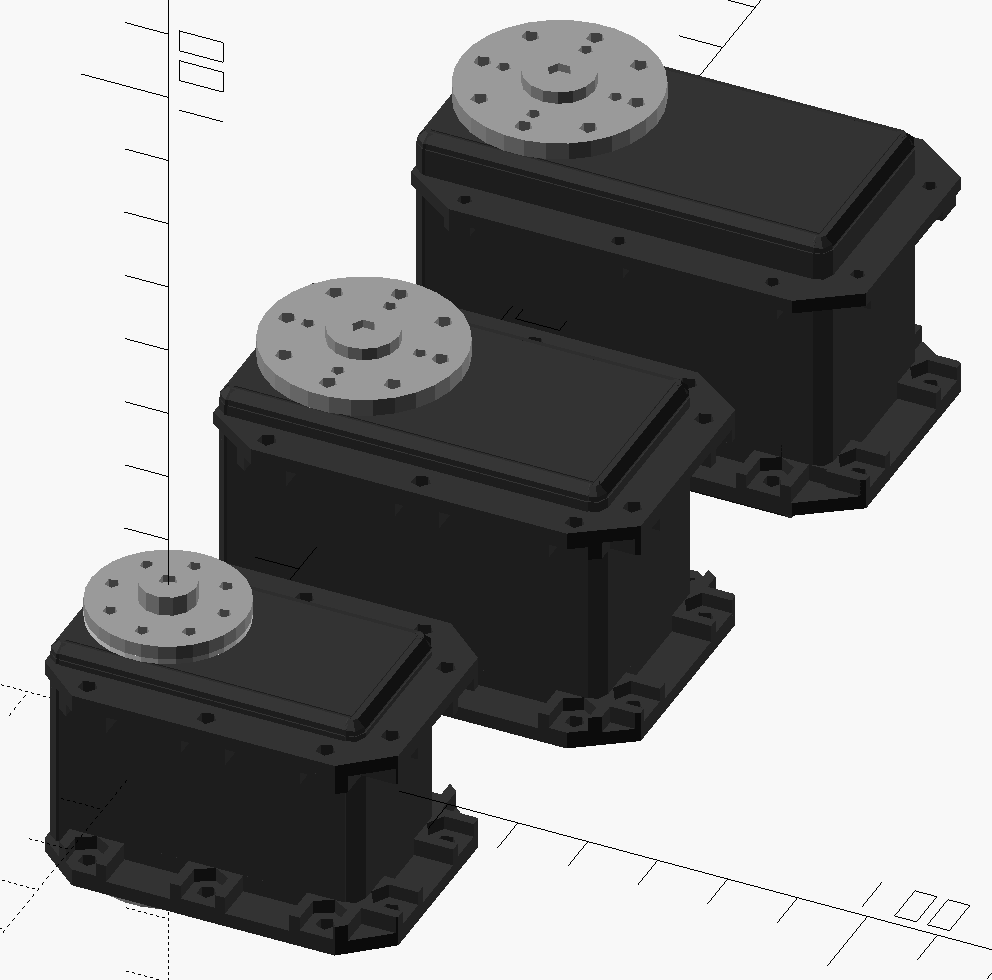}
  \caption{Depicted is the design of the thigh of the BoldBot model used in RoboCup 2018 (left).We redesigned this part with some additional support structure (middle). We will investigate whether the support for the horn and bearing will help to reduce the stress on the screws. The picture on the right shows the design of Dynamixel servos. These models can be used in OpenSCAD for designing limbs.}
  \label{fig:redesign}
\end{figure}

To help address this, we use the outer horn disc as additional support for the motors, as seen in Fig.~\ref{fig:redesign}. This gives greater support in 180 degrees (towards the model) where the parts are typically stressed. Despite being thinner, the parts are stronger as the force is spread more evenly across the model.

In the near future, we plan to investigate the use of metal inserts to further increase the strength of printed parts in the form of: small washers per screw, a larger washer inserted into the model and an embedded nut per screw.
For all our designs we utilise OpenSCAD, a tool which allows for parametric designs, enabling us to adapt the length of a limb without redesign.
Like the Darwin, all BoldBot servos are Dynamixel MX-28. It turned out that, with increasing the robot size, these servos are already too weak for the robot to stand up or to locomote. We therefore also parametrised Dynamixel servos for the models MX-28, MX-64 and MX-106\footnote{Open-sourced here: \url{https://gitlab.com/boldhearts/dynamixel-scad}} with OpenSCAD. This enables us to redesign limbs with reference to the used model more easily.

\section{Vision}
\label{sec:vision}
In previous years, our object recognition methods were based on a
lookup-table~(LUT) approach. The LUT was created based on thresholds
in HSV colour space, that were manually tweaked for each separate
competition and/or field. Besides being time consuming during setup,
the method was no longer very applicable in the modern non-colour-coded
RoboCup scenario.

The hardware upgrade that our robots received allows the application
of more advanced computer vision methods, however it is not yet
feasible to run some of the latest large-scale deep learning
models. We managed to scale such models down to be able to run fast
enough on our mobile hardware with sufficient accuracy, the full
details of which were presented at the RoboCup 2018
symposium~\cite{DijkScheunemann-18}. Here we will summarise this work.

Rather than using a direct object recognition approach, similar to the
popular YOLO and RCNN family of CNNs, which are too complex to run or
need a highly optimised domain-specific candidate selection process,
we use the more general method of semantic segmentation. Besides being
able to process a full resolution frame faster, without requiring
specific domain knowledge, this method has the additional benefit of a
single network being able to handle multiple image resolutions without
retraining. Finally, the output is a per-pixel labelling of the image,
equivalent to the output of traditional LUT based methods, so it fits
seamlessly into our existing pipeline.

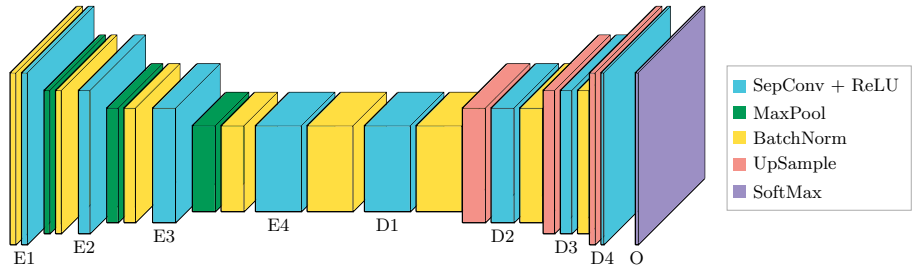
\begin{figure}
  \noindent\resizebox{\textwidth}{!}{
    \begin{tikzpicture}[baseline = (current bounding box.west)]
      
      
      \newcommand{\networkLayer}[6]{
        \def\a{#1} 
        \def\b{0.02}
        \def\c{#2} 
        \def\t{#3} 
        \ifthenelse {\equal{#6} {}} {\def\y{0}} {\def\y{#6}} 
        
        \draw[line width=0.25mm](\c+\t,0,0) -- (\c+\t,\a,0) -- (\t,\a,0);                                                      
        \draw[line width=0.25mm](\t,0,\a) -- (\c+\t,0,\a) node[midway,below] {#5} -- (\c+\t,\a,\a) -- (\t,\a,\a) -- (\t,0,\a); 
        \draw[line width=0.25mm](\c+\t,0,0) -- (\c+\t,0,\a);
        \draw[line width=0.25mm](\c+\t,\a,0) -- (\c+\t,\a,\a);
        \draw[line width=0.25mm](\t,\a,0) -- (\t,\a,\a);
        
        \filldraw[#4] (\t+\b,\b,\a) -- (\c+\t-\b,\b,\a) -- (\c+\t-\b,\a-\b,\a) -- (\t+\b,\a-\b,\a) -- (\t+\b,\b,\a); 
        \filldraw[#4] (\t+\b,\a,\a-\b) -- (\c+\t-\b,\a,\a-\b) -- (\c+\t-\b,\a,\b) -- (\t+\b,\a,\b);
        
        \ifthenelse {\equal{#4} {}}
        {} 
        {\filldraw[#4] (\c+\t,\b,\a-\b) -- (\c+\t,\b,\b) -- (\c+\t,\a-\b,\b) -- (\c+\t,\a-\b,\a-\b);} 
      }
      
      
      \networkLayer{3.0}{0.1}{-0.2}{color=Goldenrod}{}{}
      \networkLayer{3.0}{0.1}{0.0}{color=SkyBlue}{E1}{}
      \networkLayer{2.5}{0.1}{0.2}{color=ForestGreen}{}{}
      \networkLayer{2.5}{0.1}{0.4}{color=Goldenrod}{}{}
      
      \networkLayer{2.5}{0.2}{0.8}{color=SkyBlue}{E2}{}
      \networkLayer{2.0}{0.2}{1.1}{color=ForestGreen}{}{}
      \networkLayer{2.0}{0.2}{1.4}{color=Goldenrod}{}{}
      
      \networkLayer{2.0}{0.4}{1.9}{color=SkyBlue}{E3}{}
      \networkLayer{1.5}{0.4}{2.4}{color=ForestGreen}{}{}
      \networkLayer{1.5}{0.4}{2.9}{color=Goldenrod}{}{}
      
      \networkLayer{1.5}{0.8}{3.5}{color=SkyBlue}{E4}{}
      \networkLayer{1.5}{0.8}{4.4}{color=Goldenrod}{}{}
      
      \networkLayer{1.5}{0.8}{5.4}{color=SkyBlue}{D1}{}
      \networkLayer{1.5}{0.8}{6.3}{color=Goldenrod}{}{}
      
      \networkLayer{2.0}{0.4}{7.3}{color=Salmon}{}{}
      \networkLayer{2.0}{0.4}{7.8}{color=SkyBlue}{D2}{}
      \networkLayer{2.0}{0.4}{8.3}{color=Goldenrod}{}{}
      
      \networkLayer{2.5}{0.2}{8.9}{color=Salmon}{}{}
      \networkLayer{2.5}{0.2}{9.2}{color=SkyBlue}{D3}{}
      \networkLayer{2.5}{0.2}{9.5}{color=Goldenrod}{}{}
      
      \networkLayer{3.0}{0.1}{9.9}{color=Salmon}{}{}
      \networkLayer{3.0}{0.05}{10.1}{color=SkyBlue}{D4}{}
      
      \networkLayer{3.0}{0.05}{10.7}{color=Violet!50}{O}{}
    \end{tikzpicture}\hspace{2ex}
    \begin{tikzpicture}[baseline = (current bounding box.east)]
    \matrix [draw=black!33] at (current bounding box.east) {
        \node [shape=rectangle, fill=SkyBlue, label=right:SepConv + ReLU] {}; \\
        \node [shape=rectangle, fill=ForestGreen, label=right:MaxPool] {}; \\
        \node [shape=rectangle, fill=Goldenrod, label=right:BatchNorm] {}; \\
        \node [shape=rectangle, fill=Salmon, label=right:UpSample] {}; \\
        \node [shape=rectangle, fill=Violet!50, label=right:SoftMax] {};
        \\
      };
    \end{tikzpicture}
  }
  \caption{The architecture of the network consists of a series of
    fully convolutonal encoding (E1--E4) and decoding (D1--D4) steps.
    A pixelwise softmax output layer provides the final
    classifications.}\label{fig:cnn}
\end{figure}

The neural networks that we use have an ecoder-decoder structure
similar to other, large-scale, segmentation networks in the
literature, such as U-Net~\cite{ronneberger2015u} and
SegNet~\cite{badrinarayanan2017segnet}. In such networks, a first
series of convolution layers encode the input into successively lower
resolution but higher dimensional feature maps, after which a second
series of layers decode these maps into a full-resolution pixelwise
classification. This architecture is shown in
Fig.~\ref{fig:cnn}.

The main techniques that make it possible to process frames at full resolution and high rate are:
\begin{description}
\item[Depthwise Separable Convolution] By breaking up a full 3D
  convolution operation into separate per-layer 2D convolutions plus a
  1x1 depthwise convolution, the number of computations per layer is
  reduced significantly, without a great loss of performance.
\item[Stride] Using a stride of 2 reduces the number of computations in a convolution layer by 4, again without great loss (and sometimes even slight increase) of performance.
\end{description}

\begin{figure}
  \centering
  \scriptsize
  \begin{tabular}{@{}c@{}c@{}c@{\hspace{1em}}c@{}c@{}c@{}}
    \includegraphics[width=.16\textwidth]{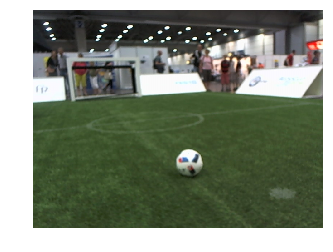} &%
    \includegraphics[width=.16\textwidth]{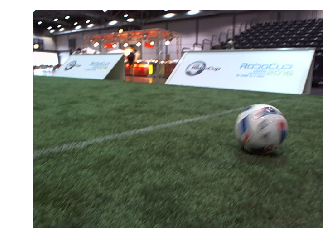} &%
    \includegraphics[width=.16\textwidth]{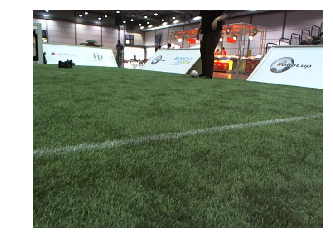} &%
    \includegraphics[width=.16\textwidth]{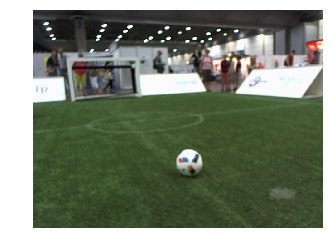} &%
    \includegraphics[width=.16\textwidth]{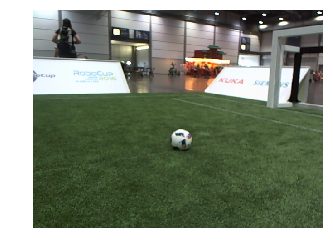} &%
    \includegraphics[width=.16\textwidth]{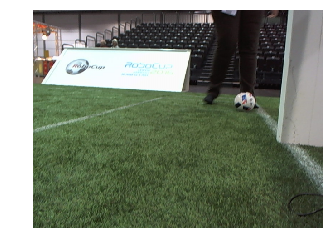} \\[-5pt]%
    \includegraphics[width=.16\textwidth]{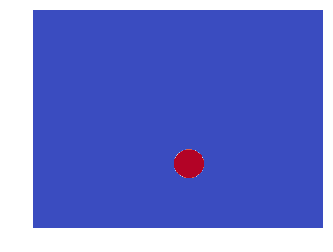} &%
    \includegraphics[width=.16\textwidth]{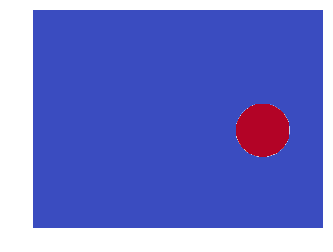} &%
    \includegraphics[width=.16\textwidth]{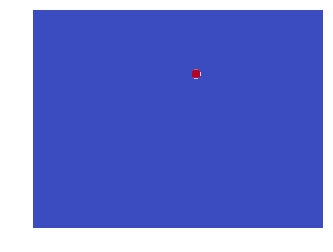} &%
    \includegraphics[width=.16\textwidth]{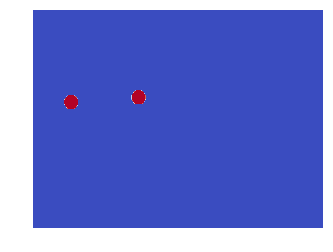} &%
    \includegraphics[width=.16\textwidth]{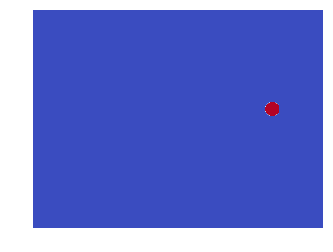} &%
    \includegraphics[width=.16\textwidth]{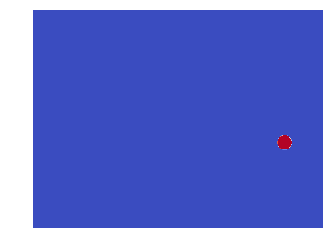} \\[-5pt]%
    \includegraphics[width=.16\textwidth]{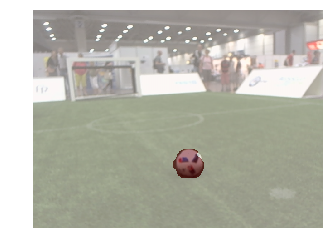} &%
    \includegraphics[width=.16\textwidth]{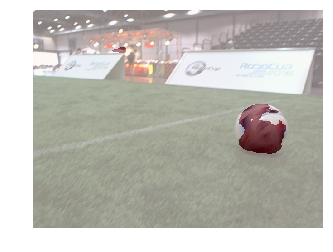} &%
    \includegraphics[width=.16\textwidth]{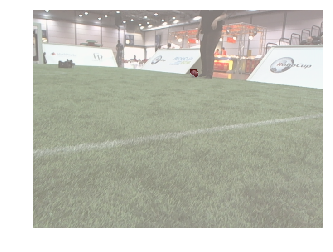} &%
    \includegraphics[width=.16\textwidth]{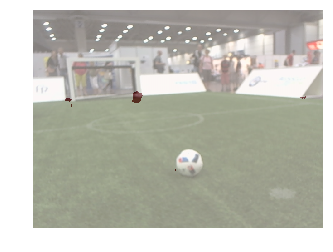} &%
    \includegraphics[width=.16\textwidth]{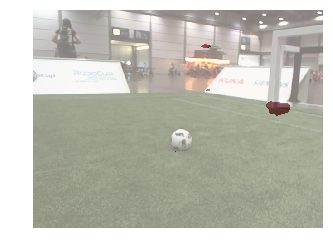} &%
    \includegraphics[width=.16\textwidth]{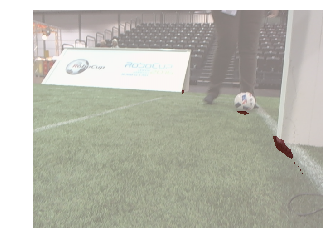} \\
  \end{tabular}
  
  \caption{Examples of input (top), target (middle) and segmentation
    outputs (bottom) of the segmentation network. Left: ball
    detection, right: goal post detection.}
  \label{fig:exampleimages}
\end{figure}

Figure~\ref{fig:exampleimages} shows typical results of these
networks. At the RoboCup 2018 competition we were able to train and run such networks
by collecting and labelling imagery at location (using the Bit-Bots'
Imagetagger\footnote{\url{https://imagetagger.bit-bots.de/}}). Training
of the network took a few hours on a laptop with a GeForce GT 750M
GPU.

During development of these methods, several open sets from the
Imagetagger were used. Additionally, we have created and released two
datasets: one with the bottom of the goalposts annotated to train goal
detection and one with ball anotations created and used during the RoboCup 2018 competition in Montreal~(See section~\ref{sec:research_contribution}).
  
\section{Using ROS 2 as Middleware}
\label{sec:middleware}

From 2013, we have developed and used our own software framework, with
all modules created from scratch, except for some that were partly
based on the source code originally supplied with the Darwin-OP
platform~\cite{DijkNoakesEtAl-14}. Although shown to be capable of performing well, over the
years the framework has become more and more complex, and being
completely custom it is now difficult for new members to get into
it. We have opted to replace the base framework completely, and have
reviewed several options: a new framework from scratch again, the
NUClear
framework\footnote{\url{https://github.com/Fastcode/NUClear}}~\cite{houliston2016nuclear},
ROS 1\footnote{\url{http://www.ros.org/}}, and
ROS 2\footnote{\url{https://index.ros.org/doc/ros2/}}.

ROS 1 was discarded as an option, as we have learned throughout the
years that efficiency of the framework is crucial for our limited
hardware, and seeing that multi-robot teams, small platforms and
real-time systems are explicit use cases that ROS 1 is not ideal for
that sparked the development of ROS 2. NUClear does offer a very
efficient, modular platform, that has been proven in the RoboCup
scenario. However, ROS 2 is currently in a state that offers most of
the same benefits, while having much wider support, including from
large entities from the industry, such as Intel and Amazon. With an eye on
the future and the transferability of skills learned by our (student)
members outside/after participation in the team, we opted to use ROS 2.

ROS 2 is based on Data Distribution Services~(DDS) for real-time systems. This connectivity framework aims for enabling scalability and real-time data exchange using a publisher-subscriber architecture.
ROS 2 sits on top of that, providing standard messages and tools to adapt DDS for robotic needs. Publishers and subscribers can be written in C++ or Python.

\begin{figure}
  \centering
  \includegraphics[width=.75\textwidth]{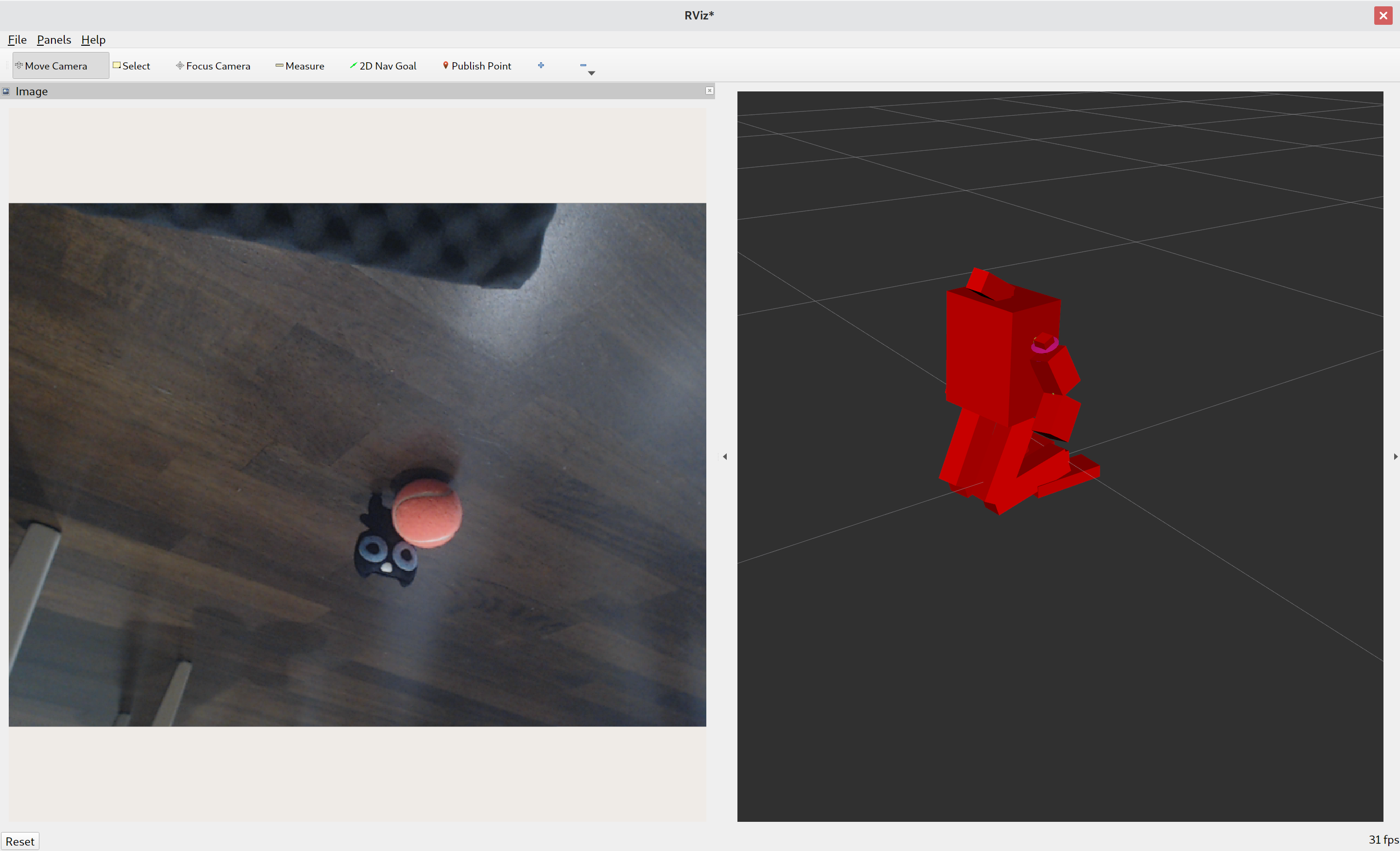}
  \caption{Screenshot of RViz2, showing camera feed and robot model built from joint states read from subcontroller.}
  \label{fig:rviz2}
\end{figure}
Our main efforts to move to use ROS 2 as our middleware consist of:
\begin{description}
\item[Creating hardware specific interfaces] Our robots are based on
  the Robotis CM-730 sub-controller. Robotis has released ROS 1 packages
  for their products, but at the moment there is no ROS 2 effort. We
  have created and are finalising a ROS 2 driver for the CM-730, which
  we will release in the near future. Figure~\ref{fig:rviz2}~(right) shows the
  result of the robot model built using the output of this driver.
\item[Porting ROS 1 packages] Only a subset of existing ROS packages
  has been ported to ROS 2. We are porting several missing
  packages, such as a USB camera
  driver, which
  we will contribute upstream (Sec.~\ref{sec:research_contribution}). Figure~\ref{fig:rviz2}~(left) shows the output
  of the camera driver.
\item[Porting our modules] Once all hardware interfaces are in place,
  we can port our existing modules over to the new platform.
\end{description}

\section{Research and Open-Source Contributions 2018}
\label{sec:research_contribution}
We presented our recent vision research at the RoboCup symposium 2018, it was ``Deep Learning for Semantic Segmentation on Minimal Hardware''~\cite{DijkScheunemann-18} and it is briefly described in section~\ref{sec:vision}.
We also published the related annotated image sets created and used during the RoboCup 2018 competition in Montreal:
\begin{itemize}
  \item goal-posts: \url{https://imagetagger.bit-bots.de/images/imageset/233}
  \item ball: \url{https://imagetagger.bit-bots.de/images/imageset/12/}
\end{itemize}%
The following is a list of open-source contributions:

\begin{itemize}
\item SCAD models for Dynamixel's MX-28, MX-64 and MX-106 servos, horns and bearings: \url{https://gitlab.com/boldhearts/dynamixel-scad}
\item porting the ROS 1 usb\_cam package to ROS 2: \url{https://github.com/ros-drivers/usb_cam/pull/106}
\item a simple, standalone 2D~simulator: \url{http://gitlab.com/boldhearts/pythocup}
\end{itemize}

\section{Acknowledgements}
\label{sec:acknowledgements}

Team Bold Hearts would like to acknowledge the crucial open source
projects used to develop our team: ROS 2, OpenSCAD, GitLab, TensorFlow and Bit-Bots Imagetagger\footnote{\url{https://www.openscad.org},
\url{https://gitlab.com/},
 \url{https://www.tensorflow.org/}, \url{https://imagetagger.bit-bots.de}}.

\end{document}